\documentclass[letterpaper, 10 pt, conference]{ieeeconf}  
\usepackage{amsmath,amsfonts}
\usepackage{algorithmic}
\usepackage{algorithm}
\usepackage{array}
\usepackage{textcomp}
\usepackage{url}
\usepackage{verbatim}
\usepackage{graphicx}
\usepackage{cite}
\usepackage{stfloats}
\usepackage{booktabs}
\usepackage{color}
\usepackage[dvipsnames]{xcolor}
\usepackage{dsfont}
\usepackage{colortbl}
\usepackage{bm}
\definecolor{MyBlue}{HTML}{1E90FF}
\usepackage[urlcolor=MyBlue, 
citecolor=MyBlue, 
linkcolor=MyBlue, 
colorlinks=true,]{hyperref}
\usepackage{multirow}
\usepackage{soul}
\usepackage{ragged2e}

\usepackage{multicol}
\usepackage{makecell}
\usepackage{booktabs}
\usepackage{amssymb}
\usepackage{bbding}
\usepackage{pifont}
\usepackage{wasysym}

\IEEEoverridecommandlockouts                              

\title{\LARGE \bf
ProxFly: Robust Control for Close Proximity Quadcopter Flight 
\\ via Residual Reinforcement Learning
}

\author{Ruiqi Zhang,
        Dingqi Zhang,
        and Mark W. Mueller
\thanks{The authors are with High Performance Robotics Lab, Department of Mechanical Engineering, University of California Berkeley, CA 94720, United States. Email: {\tt{\small \{richzhang, dingqi, mwm\}@berkeley.edu}}}%
}

\begin{document}

\maketitle
\thispagestyle{empty}
\pagestyle{empty}

\begin{abstract}
This paper proposes the ProxFly, a residual deep Reinforcement Learning (RL)-based controller for close proximity quadcopter flight. Specifically, we design a residual module on top of a cascaded controller (denoted as \textit{basic controller}) to generate high-level control commands, which compensate for external disturbances and thrust loss caused by downwash effects from other quadcopters. First, our method takes only the ego state and controllers' commands as inputs and does not rely on any communication between quadcopters, thereby reducing the bandwidth requirement. Through domain randomization, our method relaxes the requirement for accurate system identification and fine-tuned controller parameters, allowing it to adapt to changing system models. Meanwhile, our method not only reduces the proportion of unexplainable signals from the black box in control commands but also enables the RL training to skip the time-consuming exploration from scratch via guidance from the basic controller. We validate the effectiveness of the residual module in the simulation with different proximities. Moreover, we conduct the real close proximity flight test to compare ProxFly with the basic controller and an advanced model-based controller with complex aerodynamic compensation. Finally, we show that ProxFly can be used for challenging quadcopter mid-air docking, where two quadcopters fly in extreme proximity, and strong airflow significantly disrupts flight. However, our method can stabilize the quadcopter in this case and accomplish docking. The resources are available at {\tt\small\url{https://github.com/ruiqizhang99/ProxFly}}.
\end{abstract}

\section{Introduction}

Flying quadcopters in close proximity is a challenging task but has a number of real-world applications. Such scenarios arise during collaborative mapping and exploration missions, where the mission is confined to a limited workspace\cite{2017crazyswarm, 2018SRconfined}. Meanwhile, in some cases like aerial docking or payload transport\cite{2020shankardocking, karan2020flybat}, a close proximity quadcopter flight control is intended. However, as a significant challenge, the complex aerodynamic interaction occurs when quadcopters fly close to each other, and poses an additional risk and constraint for motion planning\cite{2024so2}. For example, as one quadcopter flies above another, the lower quadcopter is subjected to the downwash effect caused by the upper one. Specifically, this effect results in thrust loss, complex external forces and torques on the lower quadcopter, which is difficult to model using conventional model-based approaches\cite{karan2019aerodynamics}. Previous work reveals the high-fidelity aerodynamics modeling for quadcopter control in both free-flight\cite{karan2019aerodynamics, 2021learn2flymultiple} and over-actuated configuration\cite{2022dwaware} via the computational fluid dynamics (CFD) software and wind tunnel experiments. A good example is the flying-battery\cite{karan2020flybat}. To realize the in-air docking and charging for the flight time extension, researchers use explicit aerodynamic models for thrust compensation. However, these approaches rely on precise system identification and fine-tuning of controller parameters. Moreover, they are computationally expensive and can hardly be transferred to other diverse quadcopter models.

\begin{figure}
    \centering
    \includegraphics[width=8.6cm]{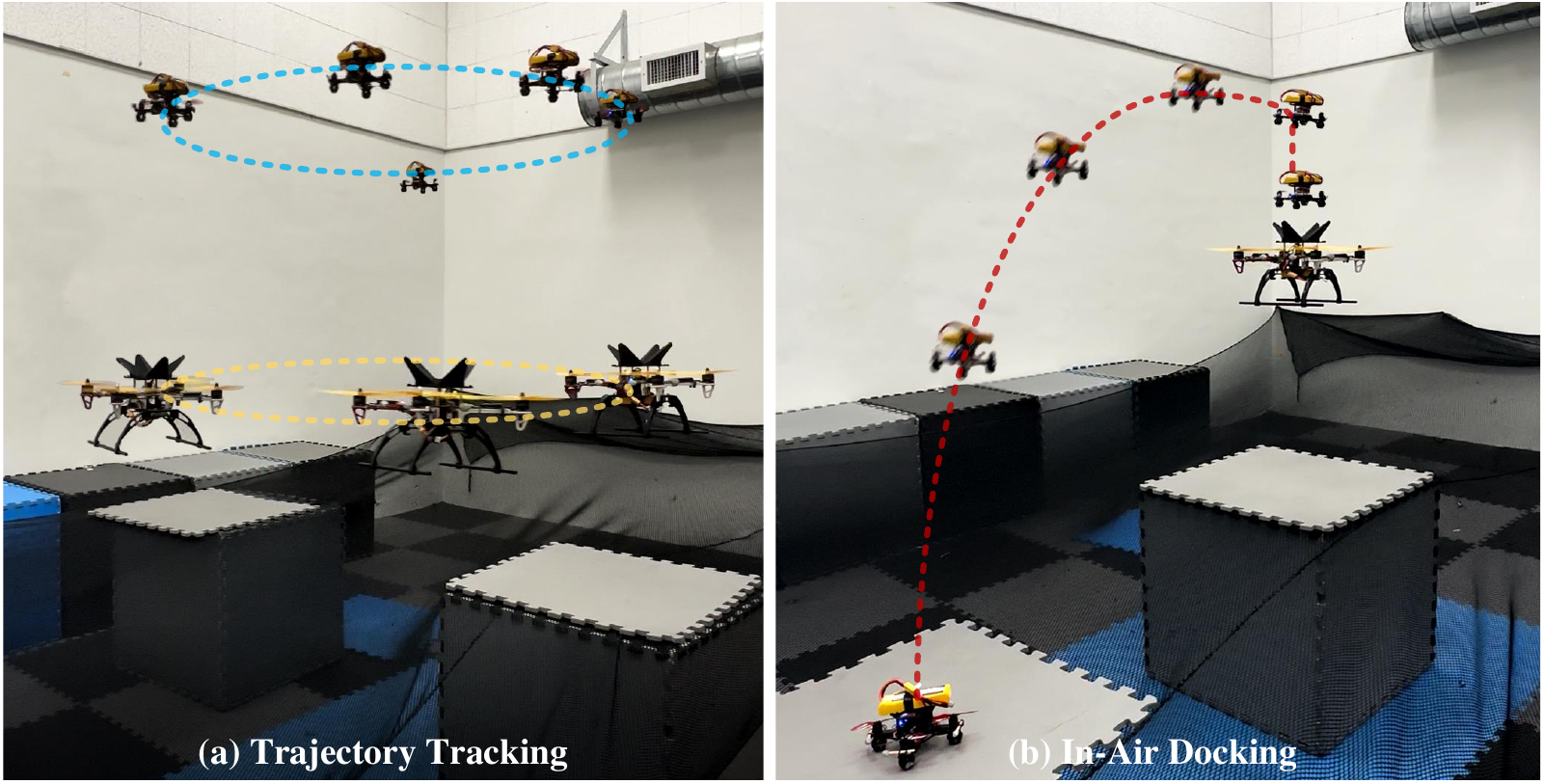}
    \caption{\textbf{Flying quadcopters in close proximity with ProxFly.} (a) Two quadcopters are tracking circular trajectories. (b) The smal quadcopter are taking off, approaching and docking with the large one in the air.}
    \label{fig:realflight}
    \vspace{-10pt}
\end{figure}

Recently, deep learning (DL) is a rapidly developing data-driven method for extracting complex latent representation from large amounts of data\cite{lecun2015deep}. It enables many challenging tasks like visual-based navigation\cite{2013dlnavigation, 2018learn2fly}, close-proximity\cite{2023dragonlearningmpc} and near-ground flight\cite{2019neurallander}, which provide new perspectives for achieving robust flight control. For example, learning a residual model to compensate for ground effects from real flight data and assist the stable control under complex aerodynamics effect\cite{2019neurallander}. However, these supervised learning methods rely on extensive data collected from human pilots and manual annotations, and the performance of learned controllers heavily depends on the skill level of the human pilots. DL can also be used to predict the downwash effect through real-world flight datasets, so that we can incorporate the learned model into motion planning, which are verified on high-speed navigation\cite{2023champdronerace}, homogeneous\cite{2020neuralswarm} and heterogeneous\cite{2022neuralswarm2} quadcopter swarms. However, these methods are only validated on micro quadcopters with weak downwash effect and large spacing, so it is uncertain whether the conclusions can be generalized to regular quadcopters and other closer proximity flight tasks.

As a powerful sequential decision-making tool, reinforcement learning (RL) can be used for many challenging tasks like flying in wind field or with off-center payload\cite{daisy2023adaptdrone, 2024learninsec, 2024e2erl4fly}. With massive data from the simulation, RL can learn a controller that maximizes cumulative returns via reward signals and avoid the risks of real flight data collection\cite{sutton2018reinforcement}. However, the policies learned through this approach tend to fit specific model configurations and tasks, so they are unsuitable for close proximity flight where system dynamics change significantly. Recent studies propose the online system identification\cite{daisy2023adaptdrone} and model-based meta-learning\cite{2021metambrl} to make quadcopters adapt to diverse external disturbances. Meanwhile, multi-agent RL control can take the state information from other robots and realize the active compensation\cite{2024marl}. However, multi-agent RL relies on the solid assumption of precise modeling and stable signal connection for communication\cite{2017mac}. Moreover, the lack of explicability in black box-based control remains a tricky and unsolved problem in the robotics community\cite{2019explainable}.

To address these challenges, we propose ProxFly, a controller for close proximity flight based on residual RL\cite{2019residualrl} (or residual policy learning\cite{silver2018rpl}). ProxFly incorporates a residual module on top of a basic cascaded controller, which generates high-level thrust and body rate commands. It compensates for external disturbances caused by the downwash effect, impulse from another quadcopter and unknown payload. Meanwhile, it takes only ego states and command outputs but doesn't rely on any forms of communication. Through the domain randomization, residual module relaxes the requirements for accurate system identification, disturbance modeling, and fine-tuning of controller parameters, so that ProxFly can adapt to diverse disturbances during the flight. Moreover, this approach reduces the proportion of unexplainable signals from the black box in the overall command and leverages guidance from a basic controller, allowing to skip time-consuming exploration from scratch of the policy network. 

In the experiment section, we first validate the effectiveness of the residual module in the simulator. After that, we conduct the comparative real flight test of ProxFly, the basic controller, and an advanced model-based controller established on the complex aerodynamics model\cite{karan2020flybat}. Specifically, we test two quadcopters for hovering and trajectory tracking in close proximity. The results show it significantly improves the basic controller's position and attitude control accuracy and achieves a comparative performance with the advanced method. Finally, we demonstrate the capability of ProxFly in quadcopter aerial docking. This is a highly challenging task since two quadcopters are required to fly at extremely close proximity, where strong downwash and local vortices disturb and reduce the propeller efficiency. Quadcopter docking also introduces impulse interference and permanently alters the quadcopter's dynamics model, which sets a high standard of adaptation and robustness for the controller.

\begin{figure}
    \centering
    \includegraphics[width=8.6cm]{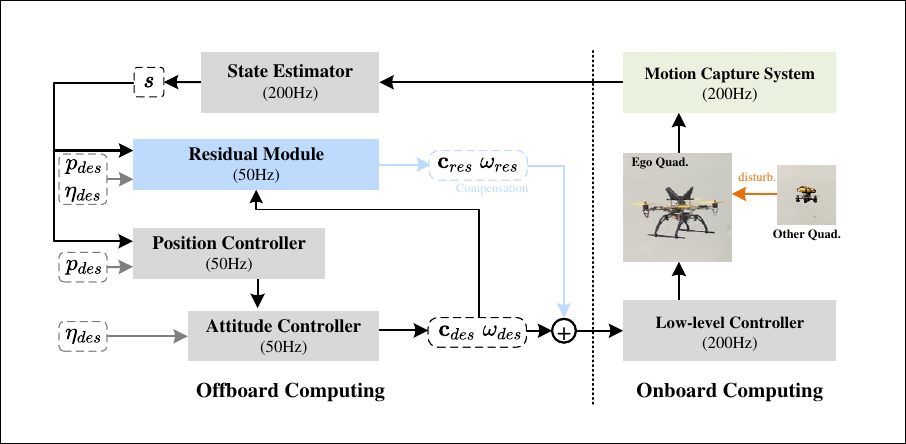}
    \caption{
        \textbf{The pipeline of ProxFly.}
        The high-level position and attitude controllers generate the desired mass-normalized thrust $\bm{c}_{des}$ and the body rate $\omega_{des}$. Meanwhile, the residual module takes the current state $s$, desired position $p_{des}$ and attitude $\eta_{des}$ and the last command from basic high-level controller. Then it generates the residual thrust $\bm{c}_{res}$ and body rates $\omega_{res}$ as a compensation of the basic controller. The overall commands can be calculated and sent to the model-based low-level controller to generate the motor speed commands. The ground truth of states are from motion capture system and the state estimator provides the estimated states to the controller.
    }
    \vspace{-10pt}
    \label{fig:pipeline}
\end{figure}

\section{Methodology}

In this section, we introduce the methodology of ProxFly. Specifically, we illustrate the principle, pipeline, setting and detailed implementation of our method.

\subsection{Residual Policy Learning}
As shown in Figure~\ref{fig:pipeline}, ProxFly uses a superposition of two control signals from the cascaded model-based controller and the residual module. The fundamental cascaded controller only requires the predefined natural frequency, damping parameters and basic model constants. It is widely used in quadcopter control and path planning\cite{karan2019aerodynamics, 2020rappids}. Although this controller design is simple and efficient, it has imperfections and the performance depends heavily on accurate model parameters. Hence, we leverage the residual RL\cite{2019residualrl} to improve the performance on the top of the cascaded controller. Specifically, the output of our ProxFly can be represented as \eqref{eq:rpl}, where the $\mathbf{u}_{cas}$ and $\bm\pi_{res}$ are the output from the cascaded controller and residual module, respectively. $\bm{s}$ and $\bm{o}$ presents the quadcopter state and observational vector. $\theta$ denotes the learnable parameter in the residual policy network. 
\begin{equation}
    \bm{\pi}(\bm{o} | \theta) = \mathbf{u}_{cas}(\bm{s}) + \bm{\pi}_{res}(\bm{o} | \theta)
    \label{eq:rpl}
\end{equation}
Note that during the training process, the gradient of policy satisfies the condition $\bigtriangledown_{\theta}\bm{\pi}(\bm{o}|\theta) = \bigtriangledown_{\theta}\bm{\pi}_{res}(\bm{o}|\theta)$, because $\mathbf{u}_{cas}$ is not a function of the inner parameter $\theta$. In other words, from the learning perspective, we can optimize the residual policy to maximize the superpose policy's reward, as their learning objectives are aligned. The advantages of this learning scheme are diverse. First, RL from scratch remains data-inefficient or intractable. However, learning a residual on top of the basic controller is simpler and skips the time-consuming exploration process at the initial stage\cite{2022rpl4ar, silver2018rpl}. Second, residual policy learning contributes to modifying its steady state error and model inaccuracy of the model-based cascaded controller and substantially improves robustness\cite{2022rpl4mrn, silver2018rpl}. Third, this method helps reduce the proportion of signals generated by the black box and facilitates the explainable result analysis of the compensatory signals based on external disturbances and the basic controller output.

In Figure~\ref{fig:pipeline}, our basic controller consists of cascaded high-level position-attitude controllers and a low-level onboard controller. The motion capture system provides ground truth position, attitude, linear velocity and body rates of the quadcopter at a frequency of $200Hz$, which is then fed back by the state estimator to the high-level controller running on the laptop. After that, it generates a four-dimensional high-level control command at $50Hz$, which includes the desired thrust and three angular velocities. Our policy network generates a residual command as compensation for the high-level control command, and then we can compute their superposition as the overall command. After that, the overall command is transmitted to the low-level controller running at $500Hz$ on the quadcopter board, which converts the high-level command into the motor speed command for execution. Additionally, we configure a simulated signal of the motion capture system and state estimator at $200Hz$. Both the basic controller and the residual policy generate commands at a frequency of $50Hz$. 

\subsection{Training Settings}
We train two independent full-enveloped flight policies for both quadcopters with the proximal policy optimization (PPO)\cite{2017ppo} algorithm. The policy and value networks are three-layer multilayer perceptrons (MLPs) with $128$ units. The first two layers use LeakyReLU\cite{2015leakyrelu} as the activation, while the last layer uses the $\tanh$ activation to normalize the output. The Adam optimizer\cite{kingma2014adam} is employed for training these networks. For each episode during the training process, we simulate $20$ seconds of flight (i.e., $10,000$ time steps at the simulation frequency of $500$). In the first $15$ seconds, the takeoff and hover phases are simulated, followed by the landing phase in the last $5$ seconds. At the beginning of each episode, the vehicle's position is randomly initialized on the ground within a $2m\times2m$ area, and we set the desired hovering position as $(0, 0, 1.2)$. During the last 5 seconds, the quadcopter is required to descend vertically at a speed of $0.2$m/s. We train the policy for $500$ epochs with $10$ episodes in each one, which can be finished in $15$ minutes with parallelized computing~\cite{hou2025spreeze} on a laptop with a Intel i7-13700H CPU. 

\subsubsection{Action and Observation Space} The policy network outputs the action $\bm{a}_{res} = \{\bm{c}_{res}, \bm{\omega}_{res} \}\sim \bm{\pi}_{res} \in \mathbb{R}^4$ that includes the residual thrust and the body rates in three directions. Empirically, we set the range of actions to $\left[-10 m\cdot s^{-2}, 10m\cdot s^{-2}\right]$ for the mass-normalized thrust and to $\left[-1rad\cdot s^{-1}, 1rad\cdot s^{-1}\right]$ for the body rates. The observation information $\bm{o}=\left\{\Delta\bm{s}, \mathbf{u}_{cas}, \bm{a}_{res} \right\} \in \mathbb{R}^{20}$ consists of the error between the desired state and current state $\Delta\bm{s} = (\bm{s}_{des}-\bm{s}) \in \mathbb{R}^{12}$, the high-level command $\mathbf{u}_{cas} \in \mathbb{R}^{4}$ from the basic cascaded controller, and the action from residual module $\bm{a}_{res}\in \mathbb{R}^4$.

\subsubsection{Reward Shaping} For a behavior trajectory sampled in the simulator for $T$ time steps, we present it as
$\tau = \left\{\bm{s}_0, \bm{a}_0, \bm{r}_0, \bm{s}_1, \bm{a}_1, \bm{r}_1,\cdots, \bm{s}_T, \bm{a}_T, \bm{r}_T \right\}$. 
In \eqref{eq:opti_obj}, the optimization objective for RL is to maximize the return of the behavior trajectory, where $p(\tau | \bm{\pi}_{\theta})$ is the likelihood of the trajectory $\tau$ under policy $\bm{\pi}_{\theta}$ and $\gamma$ is the decay rate. As shown in \eqref{eq:reward}, we design the following reward signals to encourage the quadcopter to track the position and linear velocity commands while avoiding oscillations. Empirically, we set a vector of scaling factors $\bm{\alpha} = [-1m^{-1}, -1, -0.01 Kg\cdot N^{-1}, -0.1 s\cdot rad^{-1}, 1]$ to balance different reward signals and unify them to unitless terms.
\begin{equation}
    \mathcal{J}(\bm{\pi}) = 
    \mathbb{E}_{\tau \sim p(\tau | \bm{\pi})}
    \left[ 
        \sum \nolimits _{t=0}^{T}
        \gamma^t\bm{r}_t
    \right]
    \label{eq:opti_obj}
\end{equation}
\begin{equation}
    \bm{r}_t = \bm{\alpha}\cdot
    \left[
       \bm{e}_{pos}, \bm{e}_{att}, \mathcal{P}_{\bm{c}}, \mathcal{P}_{\bm{\omega}}, \bm{r}_{\Delta t}
    \right]^T
    \label{eq:reward}
\end{equation}
\begin{itemize}
    \item \textbf{Position Error Penalty.} We use the L2-norm to set a penalty $\bm{e}_{pos} = || p_{des} - p_t ||$ for position error, where $p_{des}$ is the desired position from the planner and $p_t$ is the quadcopter position in the world frame.
    \item \textbf{Attitude Error Penalty.} We compute a unitless attitude error $\bm{e}_{att} = 3-\text{trace}(R_{t}^T\cdot R_{des}) = 2-2\cos(\omega)$, where $R_{t}^T$ and $R_{des}$ are the rotation matrix of the desired and estimated attitude, respectively.  $\text{trace}(R_{t}^T\cdot R_{des})$ is the geodesic distance on $SO(3)$ between the desire and estimated attitude. $\omega$ is the smallest rotation angle from the estimated attitude to the desired one.
    \item \textbf{Thrust Command Penalty.} Large thrust command will increase the energy consumption and command oscillation will make motors overheat. To alleviate these problems, We use the sum of two L2-norm terms $\mathcal{P}_{\bm{c}} = ||\bm{c}_{t}|| + 2\ ||\bm{c}_{t} - \bm{c}_{t-1}||$ to punish large thrust command and the oscillation of overall command. We set an empirical factor $2$ to scale the oscillation penalty.
    \item \textbf{Body Rate Command Penalty.} Similarly, we set a command penalty $\mathcal{P}_{\bm{\omega}} = ||\bm{\omega}_{t}|| + 2\ ||\bm{\omega}_{t} - \bm{\omega}_{t-1}||$ to punish large body rate commands and oscillation. The factor $2$ is set to scale the oscillation penalty as well.
    \item \textbf{Survival Reward.} At each time step, a unitless positive reward signal $\bm{r}_{\Delta t} = 0.1$ is sent to encourage any helpful action for survive and avoid risky behaviors.
\end{itemize}

\subsubsection{Randomization} To avoid the policy from overfitting to a specific configuration, we randomize the model constants of the quadcopter, as detailed in Table~\ref{tab:setup1}. We set uniform distributed random domains for the mass, inertia, and propeller efficiency. The model randomization leads to the inconsistency between the model parameters set in the basic controller and the real model, which contributes to improving the generalizability of our residual module across different quadcopter dynamics. Due to the larger thrust-weight ratio of the large quadcopter~(LQ), we set a $\pm50\%$ mass error for it while a $\pm20\%$ error for the small quadcopter~(SQ). Additionally, according to the scaling law, the size of a quadcopter correlates positively with its mass and moment of inertia but is not strictly linear. Therefore, in practice, we multiply another random factor sampled from $\mathcal{U}(0.8, 1.2)$ to the mass factor as the overall inertia factor. Mass and Inertia randomization ensures that ProxFly can maintain stable flight even if the model parameters in the basic controller are inaccurate or if a payload is added during its flight.

\renewcommand{\arraystretch}{1.1}
\begin{table}[tb]
    \centering
    \caption{The parameters of quadcopter models and domain randomization settings in the training process}
    \begin{tabular}{l|c|c}
    \toprule
    \textbf{Parameters} & \textbf{Small Quad.} & \textbf{Large Quad.} \\
    \hline
        Mass ($Kg$) & 0.280 & 0.850 \\
        Mass Factor &$\sim\mathcal{U}(0.8, 1.2)$ & $\sim\mathcal{U}(0.5, 1.5)$ \\
        Inertia around $x, y$ ($Kg\cdot m^2$)& 2.36e-4  &5.51e-3\\
        Inertia around $z$ ($Kg\cdot m^2$)& 3.03e-4  & 9.88e-3\\
        Inertia Factor & $\sim\mathcal{U}(0.8, 1.2)$ & $\sim\mathcal{U}(0.8, 1.2)$ \\
        Arm Length ($m$) & 0.058 & 0.165 \\
        Propeller Thrust Efficiency & \multirow{2}{*}{1.145e-7} & \multirow{2}{*}{7.640e-6} \\
        ($N/(rad/s)^2$) &  & \\
        Thrust Factor & $\sim\mathcal{U}(0.6, 1.2)$ & $\sim\mathcal{U}(0.6, 1.2)$\\
    \midrule
        Force Disturbance Period ($s$) & $\sim\mathcal{U}(2, 8)$ & $\sim\mathcal{U}(2, 8)$\\
        Force Disturbance & \multirow{2}{*}{$\sim\mathcal{U}(0, 0.5)$} & \multirow{2}{*}{$\sim\mathcal{U}(0, 2)$}\\
        Amplitude on $x, y$ ($N$) &  &  \\
        Force Disturbance & \multirow{2}{*}{$\sim\mathcal{U}(0, 2)$} & \multirow{2}{*}{$\sim\mathcal{U}(0, 8)$} \\
        Amplitude on $z$ ($N$) &  & \\
        Force Truncation in $x, y$ ($N$)& 0.25 & 1.00 \\
        Force Truncation in $z$ ($N$)& 1.00 & 4.00 \\
        Torque Disturbance ($N\cdot m$) & $\sim\mathcal{N}(0, 0.005)$ & $\sim\mathcal{N}(0, 0.02)$ \\
    \bottomrule
    \end{tabular}
    \vspace{-10pt}
    \label{tab:setup1}
\end{table}
For the proximity flight, the disturbance primarily comes from the downwash flow generated by the rotors. On the one hand, the downwash flow reduces the propeller efficiency of other quadcopters in the wind field, so the actual thrust decreases compared with it in the static air. On the other hand, quadcopter in wind field experiences external forces and torques acting on the body\cite{karan2019aerodynamics}. To simulate the propeller thrust loss and imbalance caused by downwash flow, we randomly initialize a propeller thrust factor from $\mathcal{U}(0.6, 1.2)$ for four propellers independently. Meanwhile, although we can calculate the force and torque disturbance with the CFD software or wind tunnel, these methods are resource-intensive, time-consuming, and cannot cover all possible flight scenarios. Therefore, we propose to present the forces in the $x, y, z$ directions as three independent triangular functions with random periods and amplitudes, as shown in Table~\ref{tab:setup1}. Since downwash flow mainly acts in the vertical direction, the force disturbance in the $z$ direction is more significant. For the torque disturbances around three axes, we simulate them using three independent Gaussian noises. This disturbance randomization gets rid of extensive randomization of aerodynamic and trajectory parameters of two quadcopter. Instead, it randomizes the external forces in the approach-hover-leave process numerically according to the conclusion in \cite{karan2019aerodynamics}. For example, variations in the approach and departure speeds of the upper quadcopter as well as the intensity of the downwash flow field affect the rate of change of external force disturbances. This is equivalent to randomizing the force disturbance period. Meanwhile, factors such as the relative hovering altitude and platform shape of the upper can determine the magnitude of the external force disturbances, we can therefore randomize the amplitude and maximum force directly. However, the shape of two drone and their relative horizontal position and yaw angle determine the torque disturbance significantly. Since our setup differs significantly from that in \cite{karan2019aerodynamics}, we simulate them with Gaussian noise instead.

\section{Experiments}

In this section, we describe our experimental setup, results, and analysis in both simulation and real-world. Overall, we validate our algorithm in simulation by demonstrating its robustness under high-fidelity simulated disturbances. Subsequently, we showcase the performance of ProxFly compared to other advanced algorithms through various real-world flight tests. At last, we demonstrate its potential in challenging aerial quadcopter docking tasks.
\subsection{Simulation Experiments and Analysis}
\begin{figure}
    \centering
    \vspace{5pt}
    \includegraphics[width=8.6cm]{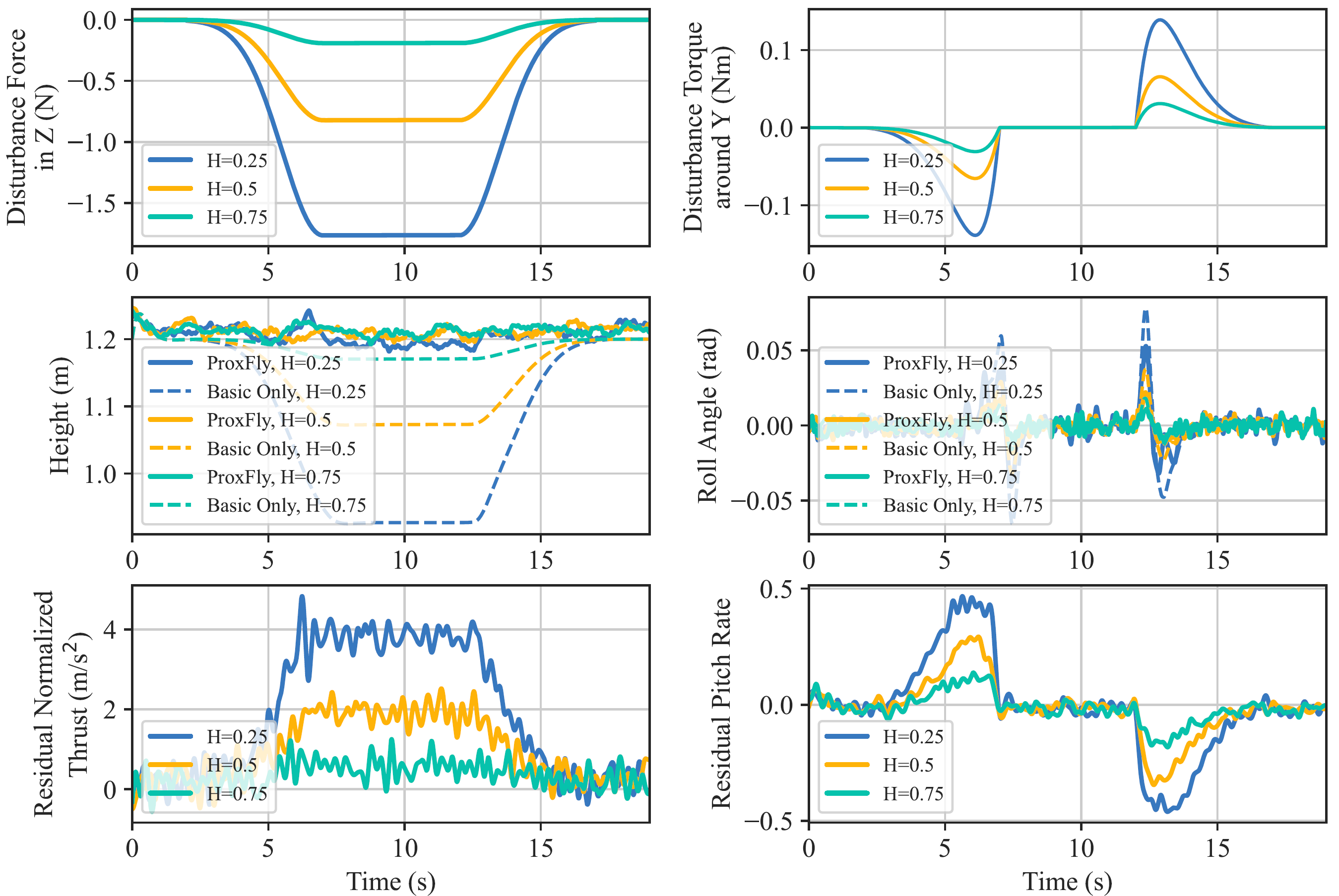}
    \caption{
        \textbf{The results of simulated experiments.}
        \textbf{First row:} The external forces in the $z$ direction and torques around the $x$ and $y$ axes from the SQ at three different height differences $H = [0.25, 0.5, 0.75]$ based on the aerodynamics model in \cite{karan2019aerodynamics}.
        \textbf{Second row:} The performance comparison of altitude, roll, and pitch attitude control using the basic controller (denoted as \textit{Basic Only}) and \textit{ProxFly}.
        \textbf{Third row:} The residual commands of mass-normalized thrust, roll rate and pitch rate.
    }
    \label{fig:sim_exp}
    \vspace{-10pt}
\end{figure}

We conduct an experiment in simulator to verify that the proposed disturbances contribute to making the residual module adaptive to high-fidelity simulated disturbances. We approximate the effect of real downwash flow as a function of the relative positions and model parameters of two quadcopters. In our setup, the LQ takes off from the ground and hovers at $(0, 0, 1.2)$, while the SQ takes off vertically from $(-1, 0, 0)$ and flies along the positive $x$-axis at a speed of $0.2$ m/s at a certain height. After hovering above the LQ for 5 seconds, the SQ continues along the positive $x$-axis and lands vertically at $(1, 0, 0)$. We set the vertical distance between the SQ and LQ as $\left\{0.25, 0.5, 0.75\right\}$ and the relative yaw angle is always $0$. Under this setup, the disturbance will be quite significant and traceable, so that we can directly apply the conclusions from \cite{karan2019aerodynamics} and reveal the corresponding compensatory behaviors generated by the residual module.

As shown in Figure~\ref{fig:sim_exp}, when the SQ approaches along the $x$-axis, the downwash force on the LQ increases exponentially as the horizontal distance between SQ and LQ decreases and reaches the peak when the SQ hovers directly above the LQ. This procedure also induces a negative torque around the $y$-axis on the SQ, which first increases and then decreases in magnitude. When the SQ leaves, the change rate of disturbances is reversed compared to the approach phase. In this process, the basic controller can correct the pitch, while it also leads to the steady-state error in altitude control, and the error becomes larger with the downwash flow increasing. On the other hand, the residual module generates thrust and body rate compensation without the relative position between the SQ and LQ.
In other words, ProxFly can correct the altitude error and reduce pitch oscillations of the basic controller with only its estimated and desired states. These behaviors are fully learned within our randomized external force and torque, which are fundamentally different from the high-fidelity simulated disturbances. However, these disturbances are derived from complex CFD simulations and wind tunnel experiments that require domain expertise. Conversely, our proposed disturbance is easy to implement.

\subsection{Real-World Experiments and Analysis}
\renewcommand{\arraystretch}{1.3}
\begin{table}[tb]
    \centering
    \caption{The results of real-world comparative experiments}
    \resizebox{\columnwidth}{!}{
    \begin{tabular}{c|c|c|c|c}
        \toprule
        \multicolumn{2}{c|}{\textbf{Tasks}} & \multirow{2}{*}{Basic Controller} & \multirow{2}{*}{FB-AeroComp\cite{karan2020flybat}} & \multirow{2}{*}{\textbf{ProxFly (Ours)}}\\
        \multicolumn{2}{c|}{\textbf{\& Metrics}} & & & \\
        \midrule
        \multirow{2}{*}{Hovering} & $\mathbf{E}_{pos}$ (m) & 0.1199 & 0.1113 & 0.0882\\
        & $\mathbf{E}_{att}$ (rad) & 0.1710 & 0.1818 & 0.0794 \\
        \hline
        Circling & $\mathbf{E}_{pos}$ (m) & 0.1867 & 0.0832 & 0.1385 \\
        (same)& $\mathbf{E}_{att}$ (rad) & 0.1976 & 0.1238 & 0.1252 \\
        \hline
        Circling & $\mathbf{E}_{pos}$ (m) & 0.1451 & 0.0983 & 0.0940\\
        (reversed)& $\mathbf{E}_{att}$ (rad) & 0.1714 & 0.0930 & 0.0996\\
        \hline
        \multirow{2}{*}{\textbf{Average}} & $\mathbf{E}_{pos}$ (m) & 0.1506 & 0.0976 & 0.1069\\
        & $\mathbf{E}_{att}$ (rad) & 0.1800 & 0.1329 & 0.1014\\
        \bottomrule
    \end{tabular}
    }
    \vspace{-10pt}
    \label{tab:real_exp}
\end{table}
\subsubsection{Comparative Experiments}
We aim to explain the performance improvements of ProxFly over the basic controller for close proximity flight control through real-world experiments. For this, we set the following tasks.
\begin{itemize}
    \item \textbf{Close Proximity Hovering.} The small quadcopter (SQ) hovers $50cm$ above the large quadcopter (LQ) for $10$ seconds. This experiment demonstrates the robustness of position and attitude control under continuous downwash flow disturbance.
    \item \textbf{Circling in the same direction.} The SQ flies $50cm$ above the LQ, and both vehicles track a circular trajectory with a diameter of $1.5m$ counterclockwise with a period of $7.5$ seconds. This experiment evaluates the controller's robustness and trajectory tracking accuracy under continuous air disturbance.
    \item \textbf{Circling in the reversed direction.} The SQ flies $50cm$ above the LQ and both vehicles track a circular trajectory with a diameter of $1.5m$ with a period of $7.5$ seconds. The SQ tracks counterclockwise while the LQ tracks clockwise. This experiment evaluates the controller's robustness and tracking accuracy under sudden air disturbance.
\end{itemize}

\textbf{Baselines.} First, as an ablation experiment, we evaluate the performance of our basic cascaded controller (denoted as \textit{Basic Controller}) in Table~\ref{tab:real_exp}, which illustrates the performance improvements brought by our residual module. Meanwhile, we compare it with the model-based fine-tuned cascaded controller\cite{karan2019aerodynamics}, which can handle complex close proximity flight tasks with the same hardware used in this paper. Technically, it first models the aerodynamics of quadcopters via both CFD software and real-world verification, and then calculates the thrust compensation from the relative position between quadcopters. Here, we denoted it as \textit{FB-AeroComp}.

\textbf{Metrics.} We compare the performance of different controllers based on the tracking accuracy of position and attitude. For position accuracy, we evaluate it by the root mean square error (RMSE) between the estimated and desired positions $\mathbf{E}_{pos}$, as shown in \eqref{eq:pos_error}. Meanwhile, similar to the reward shaping, for attitude accuracy, we use the RMSE of the minimum rotational angle $\mathbf{E}_{att}$ between the estimated and desired attitude as \eqref{eq:att_error}.
\begin{equation}
    \mathbf{E}_{pos}
    = \sqrt{\frac{1}{T}\sum \nolimits ^{T}_{t=1}(p_{des}-p_{t})^2}
    \label{eq:pos_error}
\end{equation}
\begin{equation}
    \resizebox{0.9\hsize}{!}{$
    \mathbf{E}_{att}
    = \sqrt{\frac{1}{T}\sum \nolimits ^{T}_{t=1}\left[\cos^{-1}\left(\frac{\text{trace}(R^T_t\cdot R_{des})-1}{2}\right)\right]^2}
    $}
    \label{eq:att_error}
\end{equation}
\begin{figure}
    \centering
    \vspace{5pt}
    \includegraphics[width=8.6cm]{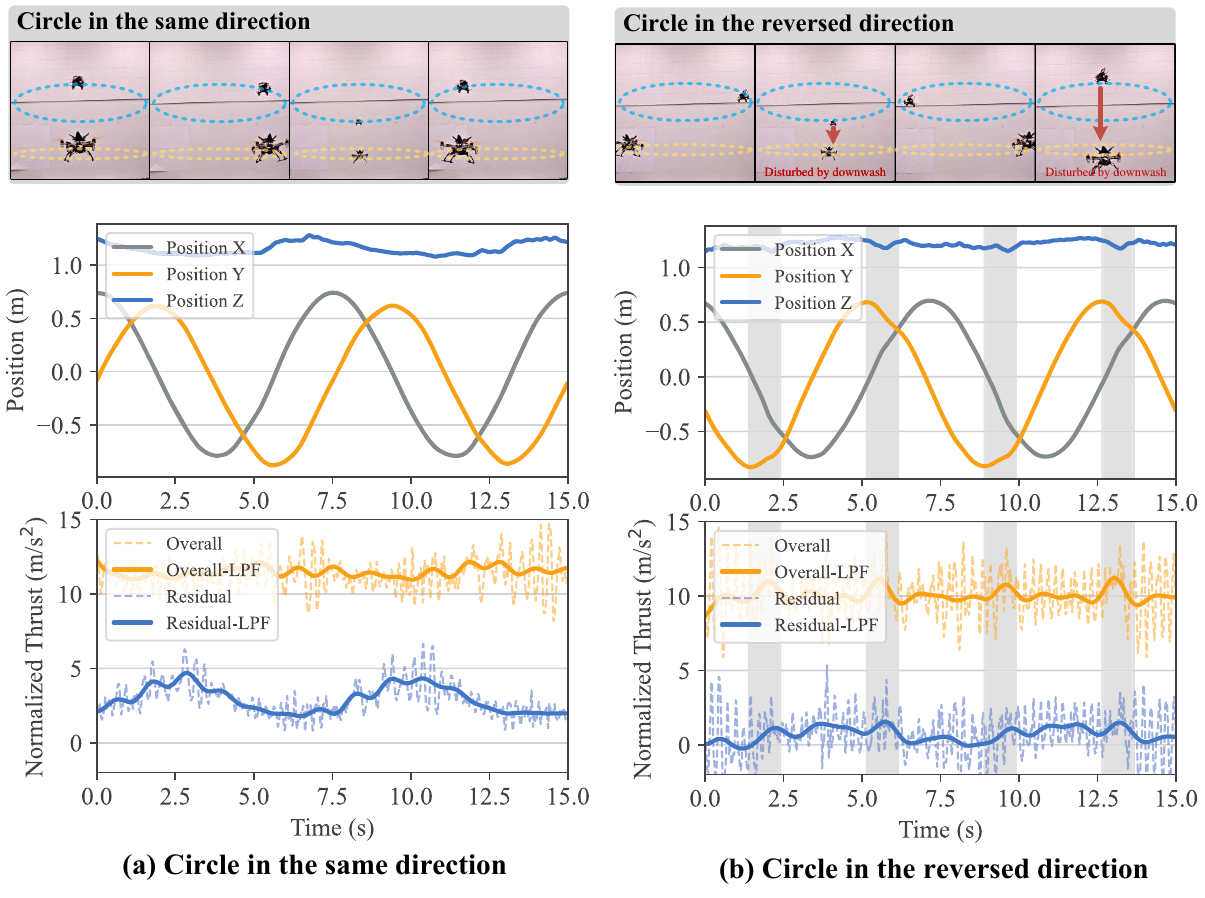}
    \caption{\textbf{The demonstration of circular trajectory tracking.}
    (a) Two quadcopters are tracking the circular trajectory counterclockwise. The residual module generates positive thrust commands to compensate the thrust loss and downward force caused by downwash flow. (b) Two quadcopters are tracking in reversed directions. When the small quadcopter is passing above the large one, the controller on the large quadcopter increases the thrust for about $1s$ for compensation. (\textit{LPF: low-pass filtered})
    }
    \vspace{-10pt}
    \label{fig:circle}
\end{figure}
\textbf{Analysis.} Compared to the basic controller, ProxFly demonstrates significant performance improvements across all three tasks. Specifically, the residual module reduces the averaged RMSE of position control by $29.0\%$ and averaged attitude error by $43.7\%$ across the three tasks. 
In the circling in the same direction and hovering tasks, these improvements result from the residual commands correcting the steady-state errors caused by continuous external aerodynamic disturbances. For the circling in the reversed direction task, the residual module allows the LQ to recover more quickly to the desired altitude after a sudden downward impulse. Meanwhile, compared to FB-AeroComp with complex aerodynamic modeling, ProxFly achieves comparable performance. In the hovering task, ProxFly reduces position error by $20.8\%$ and attitude error by $56.3\%$ compared to FB-AeroComp. For circular trajectory tracking tasks, as shown in Figure~\ref{fig:circle}, the residual module of ProxFly compensates for the downward force and thrust loss when the downwash effect happens. For example, when two quadcopters circle in the same direction, the residual module generates the lasting positive thrust command for compensation. While circling in two reversed directions, every time when the SQ flies over the LQ, the residual module generates a temporary thrust compensation.
Our method achieves the comparative performance with FB-AeroComp. Note that ProxFly is mainly trained for hovering rather than trajectory tracking. In these tasks, we implement trajectory tracking by continuously changing the hovering position, and ProxFly is highly sensitive to position and attitude errors, which leads to more aggressive actions. Meanwhile, the output commands still oscillate at a high frequency, although we have set a penalty for it during reward shaping. However, our approach does not depend on any prior assumptions of accurate models and any forms of communication between quadcopters. Essentially, ProxFly adapts to various disturbances by using a residual module to correct the behaviors from a simple controller, which simplifies the time-consuming procedure of dynamics modeling and controller fine-tuning.

\begin{figure}
    \centering
    \vspace{5pt}
    \includegraphics[width=8.6cm]{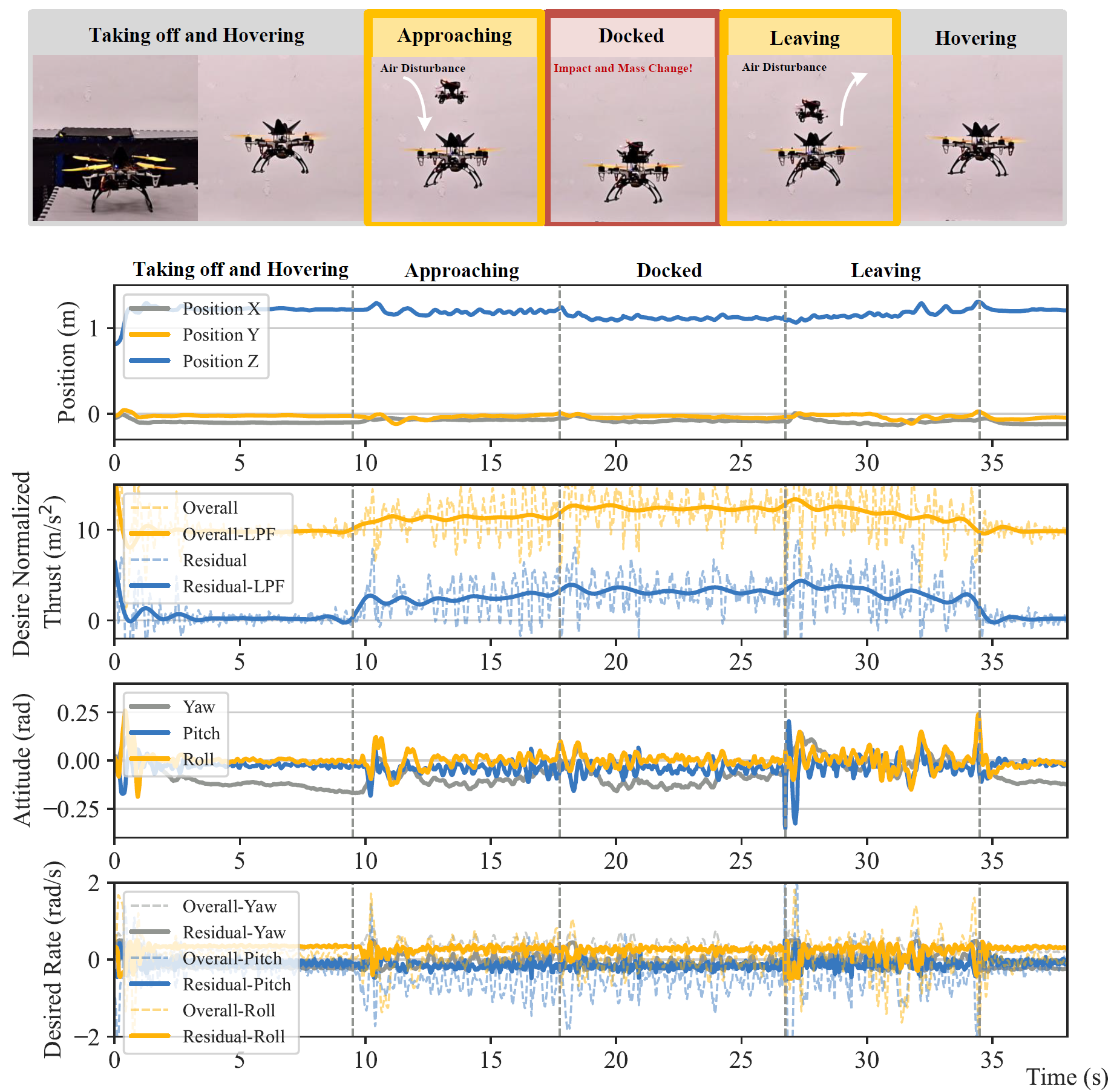}
    \caption{\textbf{The procedure of quadcopter mid-air docking.}
    \textbf{Stage 1: Taking off and hovering.} The residual RL controller only assists the basic controller on the large quadcopter (LQ) to achieve faster responses.
    \textbf{Stage 2: Small quadcopter approaching.} Small quadcopter (SQ) hovers above the LQ and vertically approaches it, which generates strong downwash flow disturbances, and the residual RL controller generates thrust and rate compensation to help stabilize LQ's position and attitude. (\textit{LPF: low-pass filtered})
    \textbf{Stage 3: Docked with LQ.} The SQ falls freely from 5cm above the LQ, gets docked with the LQ and generates an impulse. The overall mass changes and the thrust compensation from the RL controller on LQ reaches the peak.
    \textbf{Stage 4: SQ leaving.} SQ takes off from the LQ vertically, and the downwash airflow reappears and gradually decreases while the thrust compensation from the RL controller also gradually decreases.
    }
    \vspace{-10pt}
    \label{fig:docking}
\end{figure}

\subsubsection{Quadcopter Aerial Docking}

In-air charging to achieve infinite flight time can be accomplished through quadcopter docking\cite{karan2020flybat}. Due to the extreme proximity between the quadcopters, strong and complex local vortices are generated and cause severe thrust loss in both the LQ and SQ, which makes the aerodynamic modeling and control for docking highly challenging. Additionally, when the SQ lands on top of the LQ, it causes an impulse and a permanent change in dynamics. However, our ProxFly can adapt to unknown impulses and payloads and achieves robust position and attitude control. The LQ first takes off and hovers at $(0, 0, 1.2)$. Then, the SQ approaches directly above the LQ from $(0, -1, 1.7)$ and then descends vertically at a speed of $0.1$m/s until it is $10$cm above the LQ. When the horizontal distance between the two quadcopters is less than $5$cm, the SQ shuts down its motors and falls freely onto the top of the LQ, and the docking is finished. The SQ exerts an impulse on the LQ and the overall mass changes. When the undocking command is received, the SQ turns on the motors and rakes off vertically, where the thrust loss caused by downwash flow reappears.

The position, attitude, and controller outputs of the LQ are shown in Figure~\ref{fig:docking}. As the SQ approaches the LQ, the downwash flow affects the propeller's efficiency on one side of the LQ and generates a downward force and a roll torque disturbance. At this point, the residual module generates an instantaneous body rate command in the opposite direction of the roll to offset the rotation. When the SQ approaches vertically, the downward force disturbance and thrust loss acting on the LQ gradually increase. The residual module then produces a correspondingly increasing compensatory thrust command to help the LQ maintain its altitude. When the SQ gets docked with the LQ, the magnitude of the residual thrust command becomes stable. Note that the preset model parameters in the basic controller have significant errors at this point, and docking would fail if only the model-based basic controller were used. We manually send the undocking signal to the SQ via a joystick. As the undocking begins, the downwash flow and complex local vortices gradually intensify, leading to the thrust loss and imbalance of the LQ once again. At this point, the LQ experiences severe roll and pitch oscillations, and the residual module therefore generates commands opposite to the roll and pitch directions to stabilize the body. As the SQ gradually moves away from the LQ, the downward force disturbance acting on the SQ decreases. Consequently, the thrust commands generated by the residual module also gradually reduce to zero.

\section{Conclusion}
This paper proposes the ProxFly, which leverages a cascaded controller combined with residual RL for diverse close proximity tasks, including hovering, trajectory tracking, in-air docking, and undocking. This approach does not rely on any communication or accurate knowledge of system parameters but only uses a lightweight MLP to learn how to compensate for high-level commands. ProxFly leverages the strengths of both classical control and deep RL. On the one hand, it requires less domain expertise and avoids time-consuming controller parameter tuning. On the other hand, compared to end-to-end RL, it improves sampling efficiency and reduces the proportion of unexplained signals from the black box in the overall control signals. The demonstration of this paper can be found at {\tt\small\url{https://youtu.be/NhPKgzd3l6w}}.

However, ProxFly's residual module can potentially cause oscillations, which might lead to motor overheating. Meanwhile, although ProxFly relaxes the requirement of accurate modeling and is adaptive to different systems, it still needs to keep the system parameters in a reasonable range. Finally, due to hardware limitations, we only validated the performance on two regular-sized quadcopters. More types of quadcopters and experiments are needed to verify the generality of the conclusions in this paper.

\section*{Acknowledgement}
This work is supported by Hong Kong Center for Logistics Robotics. We thank William Su from the department of aerospace engineering, University of California, Berkeley for his design and modification of the quadcopter docking platform. The experimental testbed at the HiPeR  Lab is the result of contributions of many people, a full list of which can be found at {\tt\small\url{https://hiperlab.berkeley.edu/members/}}.

\bibliographystyle{ieeetr}
\bibliography{main}

\begin{thebibliography}{10}

\bibitem{2017crazyswarm}
J.~A. Preiss, W.~Honig, G.~S. Sukhatme, and N.~Ayanian, ``Crazyswarm: A large nano-quadcopter swarm,'' in {\em 2017 IEEE International Conference on Robotics and Automation (ICRA)}, pp.~3299--3304, 2017.

\bibitem{2018SRconfined}
G.~V{\'a}s{\'a}rhelyi, C.~Vir{\'a}gh, G.~Somorjai, T.~Nepusz, A.~E. Eiben, and T.~Vicsek, ``Optimized flocking of autonomous drones in confined environments,'' {\em Science Robotics}, vol.~3, no.~20, p.~eaat3536, 2018.

\bibitem{2020shankardocking}
A.~Shankar, S.~Elbaum, and C.~Detweiler, ``Dynamic path generation for multirotor aerial docking in forward flight,'' in {\em 2020 59th IEEE Conference on Decision and Control (CDC)}, pp.~1564--1571, 2020.

\bibitem{karan2020flybat}
K.~P. Jain and M.~W. Mueller, ``Flying batteries: In-flight battery switching to increase multirotor flight time,'' in {\em 2020 IEEE International Conference on Robotics and Automation (ICRA)}, pp.~3510--3516, 2020.

\bibitem{2024so2}
H.~Smith, A.~Shankar, J.~Gielis, J.~Blumenkamp, and A.~Prorok, ``So(2)-equivariant downwash models for close proximity flight,'' {\em IEEE Robotics and Automation Letters}, vol.~9, no.~2, pp.~1174--1181, 2024.

\bibitem{karan2019aerodynamics}
K.~P. Jain, T.~Fortmuller, J.~Byun, S.~A. Mäkiharju, and M.~W. Mueller, ``Modeling of aerodynamic disturbances for proximity flight of multirotors,'' in {\em 2019 International Conference on Unmanned Aircraft Systems (ICUAS)}, pp.~1261--1269, 2019.

\bibitem{2021learn2flymultiple}
J.~Panerati, H.~Zheng, S.~Zhou, J.~Xu, A.~Prorok, and A.~P. Schoellig, ``Learning to fly—a gym environment with pybullet physics for reinforcement learning of multi-agent quadcopter control,'' in {\em 2021 IEEE/RSJ International Conference on Intelligent Robots and Systems (IROS)}, pp.~7512--7519, 2021.

\bibitem{2022dwaware}
Y.~Su, C.~Chu, M.~Wang, J.~Li, L.~Yang, Y.~Zhu, and H.~Liu, ``Downwash-aware control allocation for over-actuated uav platforms,'' in {\em 2022 IEEE/RSJ International Conference on Intelligent Robots and Systems (IROS)}, pp.~10478--10485, 2022.

\bibitem{lecun2015deep}
Y.~LeCun, Y.~Bengio, and G.~Hinton, ``Deep learning,'' {\em Nature}, vol.~521, no.~7553, pp.~436--444, 2015.

\bibitem{2013dlnavigation}
S.~Ross, N.~Melik-Barkhudarov, K.~S. Shankar, A.~Wendel, D.~Dey, J.~A. Bagnell, and M.~Hebert, ``Learning monocular reactive uav control in cluttered natural environments,'' in {\em 2013 IEEE International Conference on Robotics and Automation}, pp.~1765--1772, 2013.

\bibitem{2018learn2fly}
A.~Loquercio, A.~I. Maqueda, C.~R. del Blanco, and D.~Scaramuzza, ``Dronet: Learning to fly by driving,'' {\em IEEE Robotics and Automation Letters}, vol.~3, no.~2, pp.~1088--1095, 2018.

\bibitem{2023dragonlearningmpc}
J.~Li, L.~Han, H.~Yu, Y.~Lin, Q.~Li, and Z.~Ren, ``Nonlinear mpc for quadrotors in close-proximity flight with neural network downwash prediction,'' in {\em 2023 62nd IEEE Conference on Decision and Control (CDC)}, pp.~2122--2128, 2023.

\bibitem{2019neurallander}
G.~Shi, X.~Shi, M.~O’Connell, R.~Yu, K.~Azizzadenesheli, A.~Anandkumar, Y.~Yue, and S.-J. Chung, ``Neural lander: Stable drone landing control using learned dynamics,'' in {\em 2019 International Conference on Robotics and Automation (ICRA)}, pp.~9784--9790, 2019.

\bibitem{2023champdronerace}
E.~Kaufmann, L.~Bauersfeld, A.~Loquercio, M.~M{\"u}ller, V.~Koltun, and D.~Scaramuzza, ``Champion-level drone racing using deep reinforcement learning,'' {\em Nature}, vol.~620, no.~7976, pp.~982--987, 2023.

\bibitem{2020neuralswarm}
G.~Shi, W.~Hönig, Y.~Yue, and S.-J. Chung, ``Neural-swarm: Decentralized close-proximity multirotor control using learned interactions,'' in {\em 2020 IEEE International Conference on Robotics and Automation (ICRA)}, pp.~3241--3247, 2020.

\bibitem{2022neuralswarm2}
G.~Shi, W.~Hönig, X.~Shi, Y.~Yue, and S.-J. Chung, ``Neural-swarm2: Planning and control of heterogeneous multirotor swarms using learned interactions,'' {\em IEEE Transactions on Robotics}, vol.~38, no.~2, pp.~1063--1079, 2022.

\bibitem{daisy2023adaptdrone}
D.~Zhang, A.~Loquercio, X.~Wu, A.~Kumar, J.~Malik, and M.~W. Mueller, ``Learning a single near-hover position controller for vastly different quadcopters,'' in {\em 2023 IEEE International Conference on Robotics and Automation (ICRA)}, pp.~1263--1269, 2023.

\bibitem{2024learninsec}
J.~Eschmann, D.~Albani, and G.~Loianno, ``Learning to fly in seconds,'' {\em IEEE Robotics and Automation Letters}, vol.~9, no.~7, pp.~6336--6343, 2024.

\bibitem{2024e2erl4fly}
R.~Ferede, C.~De~Wagter, D.~Izzo, and G.~C. de~Croon, ``End-to-end reinforcement learning for time-optimal quadcopter flight,'' in {\em 2024 IEEE International Conference on Robotics and Automation (ICRA)}, pp.~6172--6177, 2024.

\bibitem{sutton2018reinforcement}
R.~S. Sutton and A.~G. Barto, {\em Reinforcement learning: An introduction}.
\newblock MIT press, 2018.

\bibitem{2021metambrl}
S.~Belkhale, R.~Li, G.~Kahn, R.~McAllister, R.~Calandra, and S.~Levine, ``Model-based meta-reinforcement learning for flight with suspended payloads,'' {\em IEEE Robotics and Automation Letters}, vol.~6, no.~2, pp.~1471--1478, 2021.

\bibitem{2024marl}
R.~Zhang, J.~Hou, F.~Walter, S.~Gu, J.~Guan, F.~R{\"o}hrbein, Y.~Du, P.~Cai, G.~Chen, and A.~Knoll, ``Multi-agent reinforcement learning for autonomous driving: A survey,'' {\em arXiv preprint arXiv:2408.09675}, 2024.

\bibitem{2017mac}
J.~Qin, Q.~Ma, Y.~Shi, and L.~Wang, ``Recent advances in consensus of multi-agent systems: A brief survey,'' {\em IEEE Transactions on Industrial Electronics}, vol.~64, no.~6, pp.~4972--4983, 2017.

\bibitem{2019explainable}
B.~Beyret, A.~Shafti, and A.~A. Faisal, ``Dot-to-dot: Explainable hierarchical reinforcement learning for robotic manipulation,'' in {\em 2019 IEEE/RSJ International Conference on Intelligent Robots and Systems (IROS)}, pp.~5014--5019, 2019.

\bibitem{2019residualrl}
T.~Johannink, S.~Bahl, A.~Nair, J.~Luo, A.~Kumar, M.~Loskyll, J.~A. Ojea, E.~Solowjow, and S.~Levine, ``Residual reinforcement learning for robot control,'' in {\em 2019 International Conference on Robotics and Automation (ICRA)}, pp.~6023--6029, 2019.

\bibitem{silver2018rpl}
T.~Silver, K.~Allen, J.~Tenenbaum, and L.~Kaelbling, ``Residual policy learning,'' {\em arXiv preprint arXiv:1812.06298}, 2018.

\bibitem{2020rappids}
N.~Bucki, J.~Lee, and M.~W. Mueller, ``Rectangular pyramid partitioning using integrated depth sensors (rappids): A fast planner for multicopter navigation,'' {\em IEEE Robotics and Automation Letters}, vol.~5, no.~3, pp.~4626--4633, 2020.

\bibitem{2022rpl4ar}
R.~Zhang, J.~Hou, G.~Chen, Z.~Li, J.~Chen, and A.~Knoll, ``Residual policy learning facilitates efficient model-free autonomous racing,'' {\em IEEE Robotics and Automation Letters}, vol.~7, no.~4, pp.~11625--11632, 2022.

\bibitem{2022rpl4mrn}
R.~Zhang, G.~Chen, J.~Hou, Z.~Li, and A.~Knoll, ``Pipo: Policy optimization with permutation-invariant constraint for distributed multi-robot navigation,'' in {\em 2022 IEEE International Conference on Multisensor Fusion and Integration for Intelligent Systems (MFI)}, pp.~1--7, 2022.

\bibitem{2017ppo}
J.~Schulman, F.~Wolski, P.~Dhariwal, A.~Radford, and O.~Klimov, ``Proximal policy optimization algorithms,'' {\em arXiv preprint arXiv:1707.06347}, 2017.

\bibitem{2015leakyrelu}
B.~Xu, N.~Wang, T.~Chen, and M.~Li, ``Empirical evaluation of rectified activations in convolutional network,'' {\em arXiv preprint arXiv:1505.00853}, 2015.

\bibitem{kingma2014adam}
D.~P. Kingma, ``Adam: A method for stochastic optimization,'' {\em arXiv preprint arXiv:1412.6980}, 2014.

\bibitem{hou2025spreeze}
J.~Hou, G.~Chen, R.~Zhang, Z.~Li, S.~Gu, and C.~Jiang, ``Spreeze: High-throughput parallel reinforcement learning framework,'' {\em IEEE Transactions on Parallel and Distributed Systems}, vol.~36, no.~2, pp.~282--292, 2025.

\end{thebibliography}

\end{document}